%% file: main.tex
\documentclass[runningheads]{llncs}
\usepackage[T1]{fontenc}
\usepackage{graphicx}
\usepackage{amsmath}

\usepackage{multirow}
\usepackage{makecell}
\usepackage{arydshln}

\begin{document}
\title{The Importance of Downstream Networks in Digital Pathology Foundation Models}
\titlerunning{The Importance of Downstream Networks in Foundation Models}

\author{Gustav Bredell \and Marcel Fischer \and Przemyslaw Szostak \and Samaneh Abbasi-Sureshjani\orcidID{0000-0003-4150-6414} \and Alvaro Gomariz\orcidID{0000-0002-6172-5190}}

\authorrunning{G. Bredell et al.}

%

\institute{F. Hoffmann-La Roche AG, Basel, Switzerland}

\maketitle              

\input{sec/0_abstract}    
\input{sec/1_intro}

\input{sec/3_method}
\input{sec/4_results}
\input{sec/5_conclusion}

\bibliographystyle{splncs04}
\bibliography{main.bib}

\input{sec/X_suppl}

\end{document}

%% file: sec/0_abstract.tex
\begin{abstract}
Digital pathology has significantly advanced disease detection and pathologist efficiency through the analysis of gigapixel whole-slide images (WSI). In this process, WSIs are first divided into patches, for which a \emph{feature extractor} model is applied to obtain feature vectors, which are subsequently processed by an \emph{aggregation model} to predict the respective WSI label. With the rapid evolution of representation learning, numerous new feature extractor models, often termed foundational models, have emerged. Traditional evaluation methods rely on a static downstream aggregation model setup, encompassing a fixed architecture and hyperparameters, a practice we identify as potentially biasing the results. Our study uncovers a sensitivity of feature extractor models towards aggregation model configurations, indicating that performance comparability can be skewed based on the chosen configurations. By accounting for this sensitivity, we find that the performance of many current feature extractor models is notably similar. We support this insight by evaluating seven feature extractor models across three different datasets with 162 different aggregation model configurations. This comprehensive approach provides a more nuanced understanding of the feature extractors' sensitivity to various aggregation model configurations, leading to a fairer and more accurate assessment of new foundation models in digital pathology.

\end{abstract}

%% file: sec/1_intro.tex
\section{Introduction}
\label{sec:intro}

Digital pathology (DP) has significantly advanced with automated solutions for tasks like breast cancer~\cite{conde2022herohe} and metastases detection~\cite{litjens2018camelyon}, leveraging gigapixel whole-slide images (WSI) stained with H\&E.  The challenge of applying standard deep learning models for processing these large images has led to the adoption of the multiple instance learning (MIL) framework. In MIL, as depicted in step 1 in Figure \ref{fig:method}, WSIs are divided into patches, also known as tiles. A \emph{feature extractor} model extracts features from each tile to generate embedding vectors. These vectors, collectively referred to as a \emph{bag}, are then processed by an \emph{aggregation model} to predict the WSI label~\cite{maron1997framework,gadermayr2022multiple}. Popular choices for aggregation models include AttentionMIL~\cite{ilse2018attention} and TransMIL~\cite{shao2021transmil}, which both rely on using attention mechanisms for feature aggregation.

Beyond computational efficiency, feature extractors play a critical role in overcoming the scarcity of labeled data in DP. Using representation learning approaches feature extractors can be trained on large datasets of unlabeled images enabling their use across diverse datasets.
Since the pivotal work of Chen et al.~\cite{chen2020simple}, which significantly improved visual representations using contrastive learning (SimCLR), a range of novel representation learning approaches has been introduced. SimCLR learns representations by ensuring that the embeddings of images with the same label (positive examples) are close, whereas the embeddings of images with different labels (negative examples) are far apart. Subsquently, Grill et al.~\cite{grill2020bootstrap} showed that self-supervised learning can also be done without negative examples (BYOL). This approach was further improved and combined with transformers leading to DINO~\cite{caron2021emerging}. The most recent approaches combine masked autoencoder (MAE)~\cite{he2022masked} with self-distillation. This is the strategy used by iBOT~\cite{zhou2021ibot} and is also at the core of DINOv2~\cite{oquab2023dinov2}. 

The DP field has adapted these representation learning advancements, notably in CTransPath~\cite{wang2022transformer} and REMEDIS~\cite{azizi2023robust}. Due to the large number of tiles that can be extracted from a single WSI and the availability of large publicly available datasets, such as TCGA~\cite{tomczak2015review}, the datasets for CTransPath and REMEDIS contain 16 Mio and 50 Mio tiles, respectively. These large datasets allow the development of superior feature extractors, now commonly known as \emph{foundational models}, that generalize across datasets without the need for re-training.
Filiot et al.~\cite{filiot2023scaling} made a first step in this direction and demonstrated better classification performance compared to CTransPath by extracting feature embeddings using a model trained with iBOT. Chen et al.~\cite{chen2023general} went a step further and increased the dataset size to 100 Mio tiles while using DINOv2 to train the feature extraction model. Finally, one of the most recent feature extractors, Virchow~\cite{vorontsov2023virchow}, was also trained using the DINOv2 approach but on a dataset size of 380 Mio tiles. 

Foundational models promise to extract informative features from patches across diverse datasets. Ideally, capturing relevant features enhances downstream tasks, such as classification, while poor features hinder it. Feature extractors are often evaluated through their performance in basic classification tasks using models such as linear or K-NN classification~\cite{chen2020big}. However, in digital pathology, the use of an aggregation model to process embedding vectors and make final predictions can complicate the assessment of the feature extraction quality.  

As illustrated with an example in Figure \ref{fig:two_configs}, our analysis reveals that, whereas foundational models do indeed have some influence on the classification performance, they are highly sensitive to the aggregation model configuration. 
Thus, when comparing feature extractors, the sensitivity to the second step of the classification pipeline, namely the aggregation model, is an important variable to control for. Our contribution in this paper is twofold.
\begin{itemize} 
    \item 
    We characterize the feature extractors’ sensitivity to various aggregation model configurations, challenging traditional feature extractor evaluation methods in digital pathology.

    \item We propose a framework for stringent and fair evaluation for state-of-the-art feature extractors.
\end{itemize}

\input{figures/two_configs_vs_box}

%% file: figures/two_configs_vs_box.tex
\begin{figure}[t]
    \centering
    \includegraphics[width=.7\linewidth]{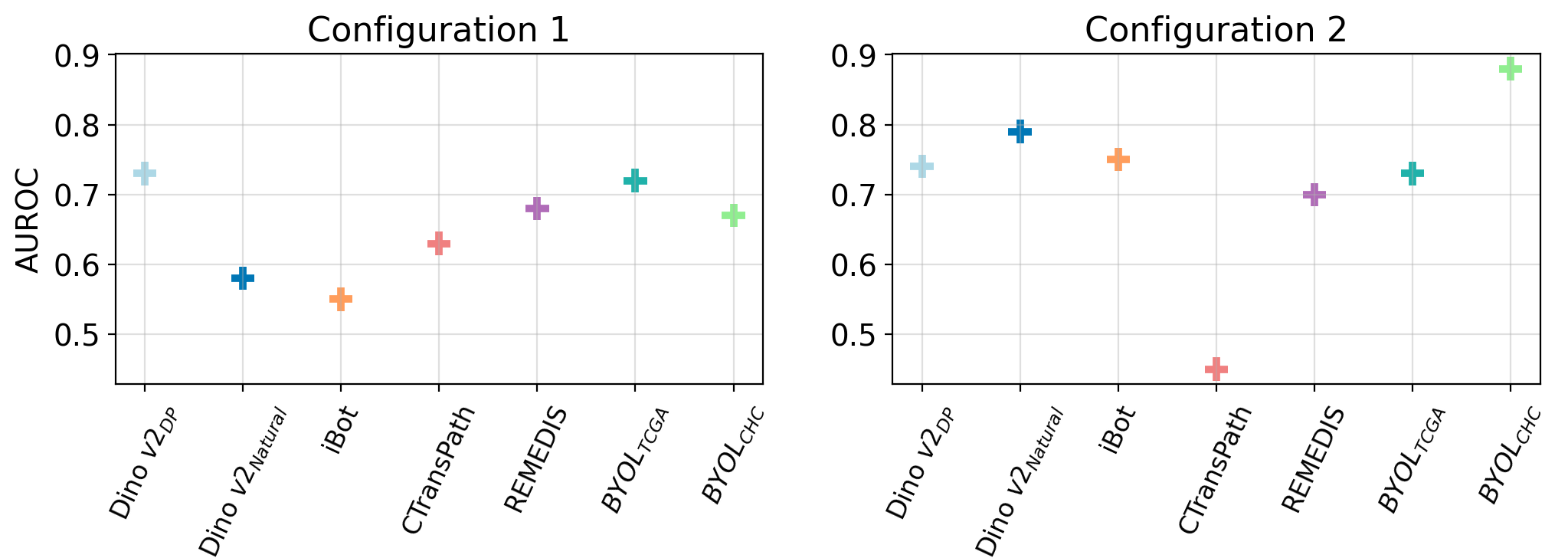}
    \caption{
Typical frameworks for evaluation of feature extraction models use fixed configurations in the aggregation models, leading to substantially different results and hence limited informative value. 
    }
    \label{fig:two_configs}
\end{figure}

%% file: sec/3_method.tex
\section{Experimental Setup and Methods}
\label{sec:method}
This section outlines the classification pipeline for whole-slide images (WSIs) and outlines the framework we use for evaluating the sensitivity of feature extractors towards aggregation model configurations.

\subsection{Pipeline for classification of WSIs}
As depicted in Figure \ref{fig:method}, the typical MIL pipeline for DP requires two models to obtain a final classification for a given WSI. 
First, a feature extraction model leverages recent self-supervised learning advancements and extensive datasets to generalize across tasks and datasets~\cite{bommasani2021opportunities,vorontsov2023virchow,chen2023general}. This model is applied to tiles in a WSI to produce feature embedding vectors. Next, a smaller aggregation model, specific to each dataset, processes the extracted embeddings to aggregate information and classify the WSI. In contrast to the feature extraction model, this aggregation model is re-trained for each dataset. 

\input{figures/method}
Feature extraction models have been compared under a single aggregation model configuration, i.e. fixed model architecture and hyperparameters~\cite{chen2023general,kang2023benchmarking,azizi2023robust}.
Figure~\ref{fig:two_configs} illustrates the significant impact of aggregation model configuration choice on performance, rendering widely adopted evaluation frameworks suboptimal. Indeed, fixed aggregation model configurations can favour some feature extraction models while penalizing others. We outline an experimental setup to thoroughly address two critical questions:

\noindent{\textbf{Question 1:}} Can a single aggregation model configuration optimally support various feature extraction models?

\noindent{\textbf{Question 2:}}  How do state-of-the-art feature extraction models perform relative to each other when controlling for different aggregation model configurations?

\subsection{Feature Extraction Models}
To explore our research questions, we assess seven feature extraction models, with details and characteristics provided in Supplementary Table~S1.

We begin by evaluating the ViT-L model from DINOv2, trained on 142 million natural images with 300 million parameters, to assess the applicability of models trained on natural images for DP~\cite{oquab2023dinov2}. Additionally, we explore recently published models specifically designed for DP: CTransPath~\cite{wang2022transformer}, REMEDIS~\cite{azizi2023robust}, and iBOT~\cite{filiot2023scaling} (teacher model), all trained on extensive DP datasets as detailed in Supplementary Table~S1. Lastly, we investigate three feature extracting models trained in-house. 

We train from scratch a DINOv2 ViT-L model on TCGA and a vast in-house dataset with diverse tissue types and real-world data. The total training dataset size is 35 million $224\times244$ tiles. WSIs are usually acquired at different magnifications. Tiles from 20$\times$ magnification offer broader content, while 40$\times$ magnification tiles provide finer tissue details due to their higher resolution. We train at both 20$\times$ and 40$\times$ magnifications to capture different features, following successful strategies in the literature~\cite{kang2023benchmarking}. We employ the official DINOv2 repository with the the default parameters for ViT-L/16 training, with a few exceptions. We decrease the batch size to 352 due to computational constraints. Due to the decreased batch size, we increase the number of epochs to 270, warm-up epochs to 25 and adjust the learning rate to $1.375\times10^{-3}$ according to the heuristic by Goyal et al.~\cite{goyal2017accurate}. 

Two ResNet-50 models are trained using BYOL: The first, $\text{BYOL}_{\text{TCGA}}$, utilizes 2 million tiles randomly sampled from the TCGA dataset at 20$\times$ magnification, featuring a smaller dataset and model size for comparison. The second, $\text{BYOL}_{\text{CHC}}$ (according to the first letter of each of the three evaluation datasets), is also trained on 2 million tiles but randomly sampled from the training set of the evaluation datasets. Thus even though the training dataset is small compared to the other published models, there is no domain gap between the dataset on which it is trained and evaluated on. This strategy ensures direct relevance to the evaluated datasets, potentially offsetting the smaller scale of the model with its domain specificity.

\subsection{Aggregation Model Configurations}
To investigate the performance fluctuation of the MIL pipeline when the feature extraction model is fixed and the aggregation model configuration change, we use different network hyperparameters and two well adopted aggregation model architectures: AttentionMIL~\cite{ilse2018attention}, which uses an attention mechanism to aggregate tile information and assumes no interdependency between the tiles. TransMIL~\cite{shao2021transmil}, which learns inter-tile dependencies by using the self-attention mechanism of transformers, in particular the Nyströmformer~\cite{xiong2021nystromformer}. 
We change four hyperparameters with three distinct values each as shown in Table~\ref{tab:hyperparameters}. These are decided heuristically with preliminary experiments assessing the influence and effective range of each hyperparameter. The resulting 162 different configurations (81 for each of the 2 architectures) are outlined below.
\input{tables/hyperparameters}

\textbf{AttentionMIL:} When creating a batch for the aggregation model during training, there are two relevant parameters. One is the \emph{bag size}, which determines the amount of tiles that is sampled from a particular WSI. The second is the \emph{bags per batch}, determining how many bags from different WSIs are collected to form a batch. Here, we vary only the bag size parameter since it showed a larger influence. The final batch size=\emph{bag size}*\emph{bags per batch}. The \emph{Layers} parameter corresponds to the number of nodes in the fully connected (FC) layers in the aggregation model. The list of numbers in Table~\ref{tab:hyperparameters} indicate the number of nodes for each layer. Lastly, the dropout parameter refers to the dropout which is applied at every layer of the aggregation model.

\textbf{TransMIL:} The selected hyperparameters for TransMIL are different due to the model architecture being a transformer, which does not employ FC layers. \emph{Layers} refers to the number of Nyströmformer attention blocks. We also reduce the maximal bag size to 2048 due to computational limitations. 

Both models share fixed training parameters: a weight decay of $10^{-5}$, four \emph{bags per batch}, AdamW optimizer, weighted cross entropy loss, and a cosine annealing scheduler. Aggregation models are trained for 50 epochs to ensure convergence within our configuration range.

\subsection{Evaluation Datasets}
Our study evaluates binary classification performance of feature extraction models across three distinct DP datasets. 
Thereby providing a more generalizable answer to our research questions. These datasets, comprising H\&E-stained histopathology slides WSIs, allow us to assess each feature extractor under 162 different aggregation model configurations. This comprehensive approach, covering 7 feature extractors, 162 aggregation model configurations, and 3 datasets, culminates in a total comparison of $7\times162\times3=3402$ experimental configurations.

Performance metrics include the area under the receiver operating characteristic curve (AUROC) and average precision (AP), both ranging from 0 to 1. A higher AUROC indicates superior distinction between the positive and negative classes, while a higher AP reflects more accurate predictions of positive instances across all recall levels, effectively balancing precision and recall. Performance metrics are derived from the test set, using the aggregation model's epoch with best validation score during training.

\textbf{COO:} Binary classification of cell of origin (COO).  Each image contains the COO prediction label of activated B-cell like (ABC) or germinal center B-cell like (GCB) tumors in diffuse large B-cell lymphoma (DLBCL). 709 WSIs from two internal datasets were used. This data closely mirrors real-world data, since it is crucial to assess classification approaches in DP using such data and tasks. The WSIs (40$\times$ magnification) have been scanned by Ventana DP200 scanners. The artifact-free tissue tiles of this dataset were combined and randomly split into 70\% training set, 15\% validation set and 15\% test set.

\textbf{Camelyon16:} Binary classification of cancer metastases vs. healthy in H\&E images of lymph node tissue. The Camelyon16 dataset~\cite{litjens2018camelyon} consists of 400 WSIs of sentinel lymph nodes. The dataset is publicly available. For our evaluation, all artifact-free tissue tiles were used as well as the official train-test split. 20\% of the training data was used as the validation set. 

\textbf{Herohe:} Binary classification of breast cancer human epidermal growth factor receptor 2 (HER2) using the publicly available Herohe~\cite{conde2022herohe} dataset. Each H\&E stained WSI is either labeled as HER2 positive or HER2 negative. We use the artifact-free tiles from tumor regions detected with an in-house tumor segmentation model. The 508 WSIs are split according to the official train-test split. 20\% of the training data is used as the validation set.

%% file: figures/method.tex
\begin{figure*}[ht]
    \centering
    \includegraphics[width=\textwidth]{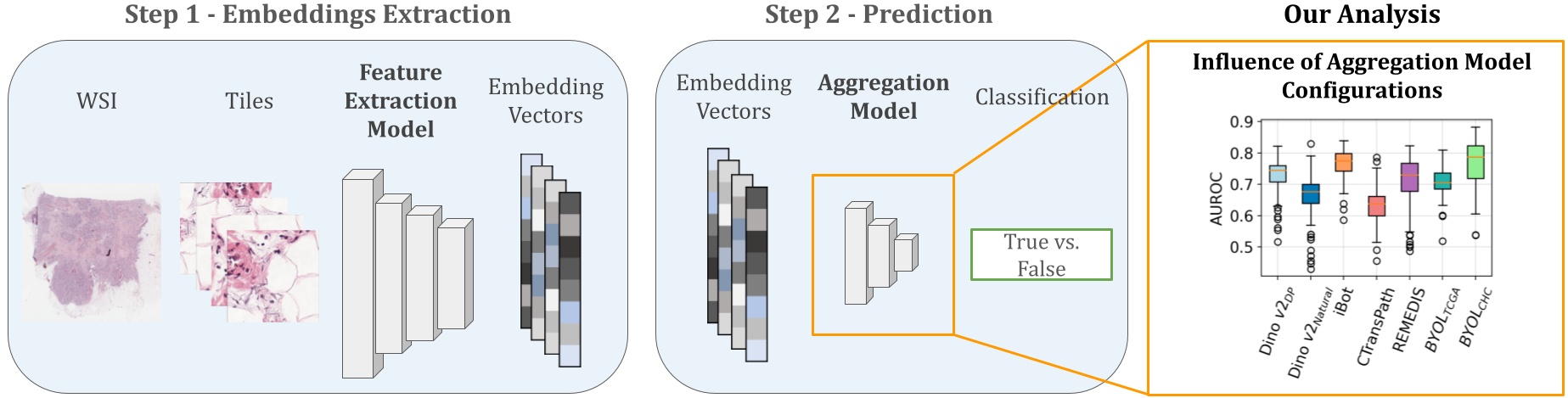}
    \caption{Illustration of the typical classification pipeline with MIL in digital pathology. 
    }
    \label{fig:method}
\end{figure*}

%% file: tables/hyperparameters.tex
\begin{table}[htb]
    \centering
    \caption{Set of hyperparameter values for each aggregation model. Layers refer to fully connected layers in AttentionMIL and to attention blocks in TransMIL.}
    \begin{tabular}{|l|c|c|}
        \hline
        \textbf{Hyperpar.} & \textbf{AttentionMIL} & \textbf{TransMIL} \\
        \hline
        Learn. rate & 1e-4, 1e-3, 1e-2 & 1e-5, 1e-4, 1e-3\\
        \hline
        Bag size & 128, 1024, 8192 & 128, 1024, 2048\\
        \hline
        Layers & (512), (512, 384, 384), & 1, 2, 3 \\
         &  (512, 256, 128, 64, 32) & \\
        \hline
        Dropout & 0.00, 0.25, 0.50 & 0.00, 0.25, 0.50\\
        \hline
    \end{tabular}
    \label{tab:hyperparameters}
    \vspace{-10pt}
\end{table}

%% file: sec/4_results.tex
\section{Results}
\label{sec:results}
This section addresses our initial inquiries, first assessing if a universally optimal aggregation model configuration exists for multiple feature extraction (foundation) models, and then comparing different feature extractors considering performance variability across aggregation model setups.

\subsection{Aggregation Model Configuration Influence}
Our analysis begins by evaluating the sensitivity of feature extractors, or foundation models, to various aggregation model configurations. 

The heatmap in Figure \ref{fig:results_heatmap} displays the classification performance across all aggregation model configurations. Trends are consistent across both AUROC and AP scores. The heatmap legend aids in identifying patterns, such as configurations with the lowest learning rate positioned on the left of each feature aggregator, which tend to yield lower performance in the COO dataset when using CTransPath but not for other features extractors. Analysis of these heatmaps reveals:
\input{figures/results_heatmap}

\noindent{\textbf{Lack of a universal configuration:}} No single aggregation model configuration consistently outperforms across all feature extractors, as indicated by the absence of a uniformly bright column across models.

\noindent{\textbf{Dataset-specific configurations:}} Optimal configurations for a given feature extractor vary by dataset. While certain parameters like learning rate for AttentionMIL and the number of attention blocks for TransMIL show some dataset-specific importance, no definitive pattern emerges across datasets or models, suggesting the need for investigation of model-specific configurations.

These results highlight the need for evaluating a diverse range of configurations in the aggregation model. This approach would ascertain that any observed superiority of one feature extraction model over another is not simply attributed to the specific aggregation model setup selected.

\subsection{Feature Extractor Comparison}
Figure \ref{fig:results_mil_boxplots} diverges from the standard practice of showing a single outcome for a fixed aggregation model setup by presenting feature extractor model performance across all 162 configurations for various datasets. Through box plots, we observe substantial performance overlap among feature extraction models despite the variance across configurations. Key insights include:
\input{figures/results_mil_boxplots}

\noindent{\textbf{Training on DP datasets is necessary}}: The DINOv2 model trained on natural images performs poorly compared to all other models, consistently for all the datasets. 

\noindent{\textbf{Comparable Performance Across Model Sizes}}: The relatively small model $\text{BYOL}_{\text{TCGA}}$ matches the performance of larger ones, suggesting that larger models are not necessarily better for DP. This echoes Filiot et al.'s~\cite{filiot2023scaling} findings that a ViT-B model can outperform a ViT-L model.

\noindent{\textbf{Feature extractors generalize well}}:  The $\text{BYOL}_{\text{CHC}}$ model, trained on WSIs from evaluation datasets, shows good performance across all datasets as expected. Interestingly, its performance is not much higher than that of other models such as $\text{DINOv2}_{\text{DP}}$, iBOT and $\text{BYOL}_{\text{TCGA}}$. This observation confirms that the feature extraction models have the capability to generalize well.

%% file: figures/results_heatmap.tex
\begin{figure}[t]
    \centering
    \includegraphics[width=0.9\linewidth]
    {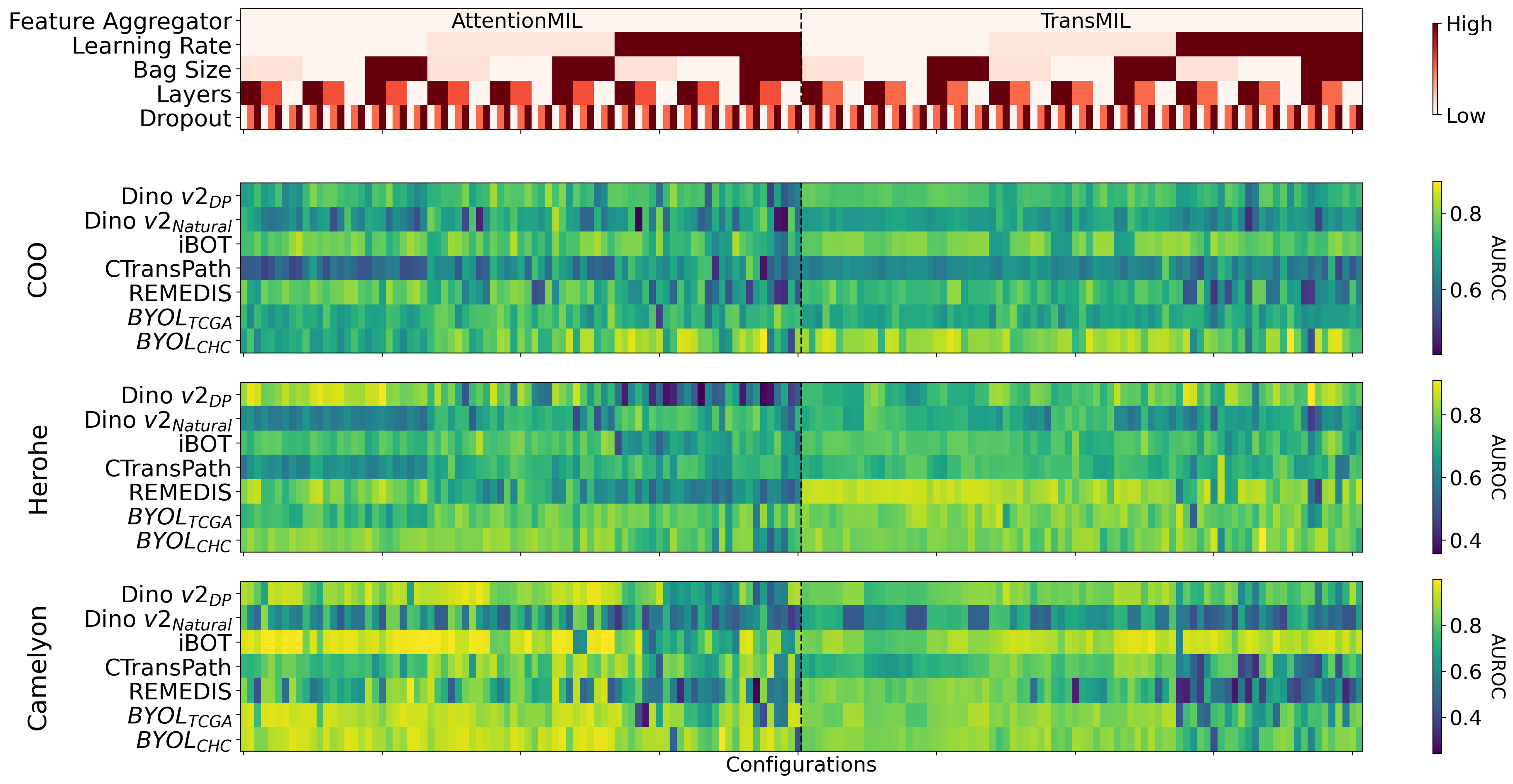}
    \caption{
    The heatmap shows the performance of every aggregation model configuration set for each feature extraction model. The red colored legend shows how the configurations are ordered on the heatmap.
    }
    \label{fig:results_heatmap}
\end{figure}

%% file: figures/results_mil_boxplots.tex
\begin{figure}[t]
    \centering
    \includegraphics[width=0.9\textwidth]{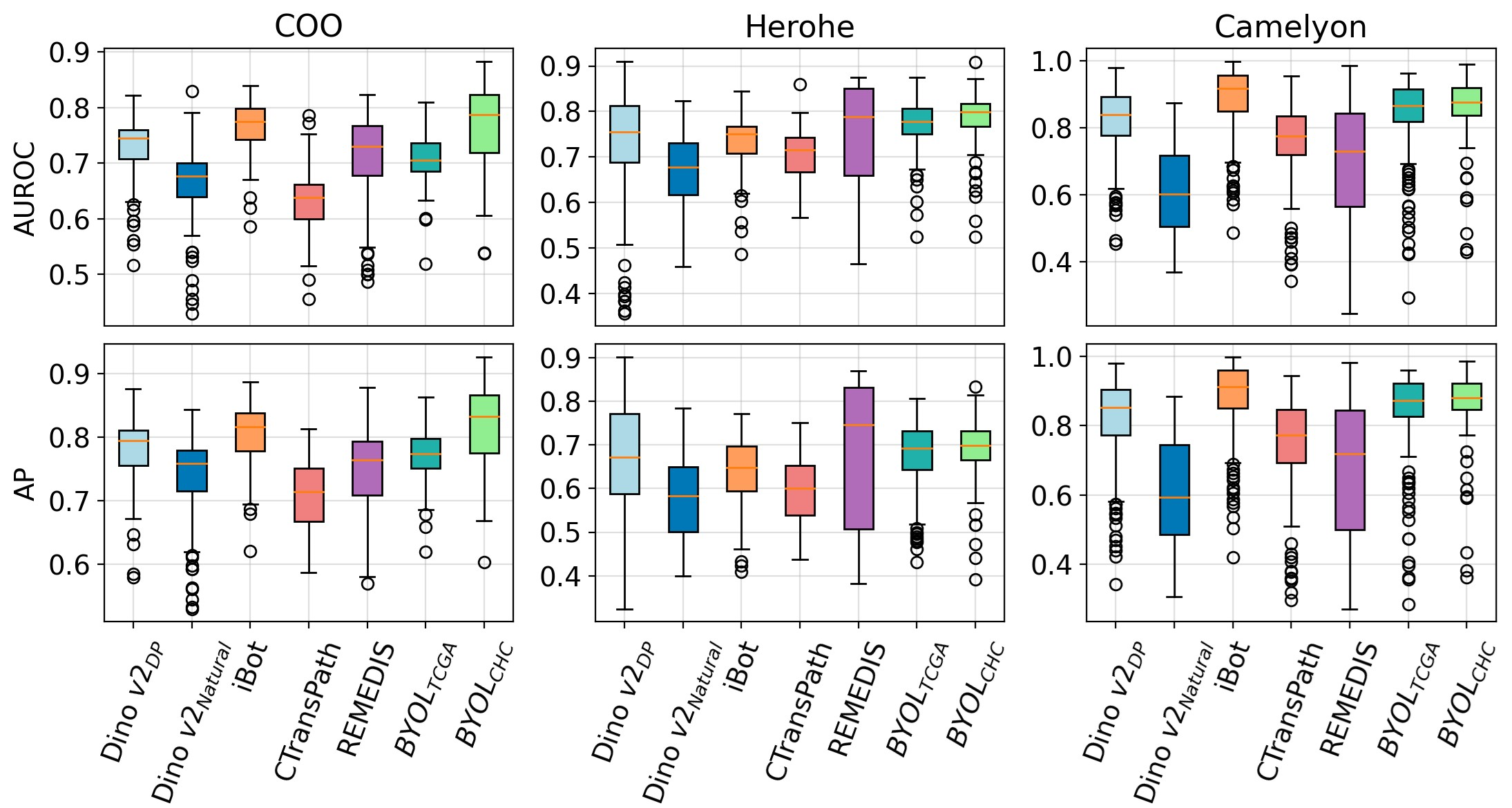}
    \caption{Comparison of 7 feature extraction models across 162 different aggregation model configurations, which include 2 architectures with 81 parameters each.
    }
    \label{fig:results_mil_boxplots}
\end{figure}

%% file: sec/5_conclusion.tex
\section{Conclusion}

In this study, we challenge the prevailing methodology for comparing foundation models in digital pathology literature, demonstrating that it may yield misleading results.
We show that due to the high sensitivity of feature extraction models to downstream aggregation model configurations, relying solely on a single aggregation model configuration can disproportionately favor certain feature extractor models while disadvantaging others. Hence, we propose evaluating foundation models across different configurations for fairer comparisons.
Our comprehensive analysis, taking into account performance variations across multiple configurations of the aggregation model, reveals a considerable overlap in performance between different foundation or feature extractor models. Significantly, we find no universal aggregation model configuration that is uniformly effective for all feature extractors.
Our work is limited though by only looking at classification tasks. In addition, the DINOv2 model we trained on digital pathology images might be subpar to other models due to computational and dataset limitations. Nevertheless, we believe this work will contribute to a more nuanced evaluation of foundation models that will help gain insight and further accelerate this rapidly evolving field.

%% file: sec/X_suppl.tex
\clearpage

\newcommand{\beginsupplement}{%
        \setcounter{table}{0}
        \renewcommand{\thetable}{S\arabic{table}}%
        \setcounter{figure}{0}
        \renewcommand{\thefigure}{S\arabic{figure}}%
     }
     
\beginsupplement

\Large {\bf{Supplementary Material\\}}

\input{tables/tab1_feature_extractor_comparison}
\input{figures/supp_results_heatmap_AP}

\input{tables/tab_results_supp}

%% file: tables/tab1_feature_extractor_comparison.tex
\begin{table*}[h]
    \centering
    \caption{List of the feature extraction models used for evaluation with some of their key properties.}
    \label{tab:feature_extraction}
    \resizebox{\linewidth}{!}{
    {\renewcommand{\arraystretch}{1.4}
    \begin{tabular}{|c|c|c|c|c|c|c|c|c|}
    \cline{2-9}
   \multicolumn{1}{c|}{}
   & \multirow{2}{*}{\textbf{Learning Approach}}
   & \multirow{2}{*}{\textbf{Model Details}}
   & \multirow{2}{*}{\makecell{\textbf{Model Size}\\ \textbf{(Mio.)}}}
   & \multirow{2}{*}{\makecell{\textbf{Embed.}\\ \textbf{Size}}}
   & \multirow{2}{*}{\makecell{\textbf{Dataset}\\ \textbf{Content}}}
   & \multicolumn{2}{c|}{\textbf{Dataset Size}} 
   & \multirow{2}{*}{\makecell{\textbf{In-house}\\ \textbf{training}}} \\ \cline{7-8}
   \multicolumn{1}{c|}{} 
   & & & & & & \textbf{Tiles (x$10^6$)} & \textbf{WSIs (x$10^3$)} &\\ \hline
   
\hline

    \textbf{\begin{tabular}[c]{@{}c@{}}Dino v2\\ (DP)\end{tabular}} & Combination & ViT-L/16 & 300 & 1024 & \begin{tabular}[c]{@{}c@{}}TCGA + in-house\end{tabular} & 35 & 70 & yes \\ \hline
    \textbf{\begin{tabular}[c]{@{}c@{}}Dino v2\\ (Natural)\end{tabular}} & Combination & ViT-L/14 distilled & 300 & 1024 & \begin{tabular}[c]{@{}c@{}}LVD-142M\\ (natural images)\end{tabular} & 142 & N/A & no \\ \hline
    \textbf{iBOT} & Combination & ViT-B & 86 & 768 & TCGA & 43 & 6 & no \\ \hline
    \textbf{CTransPath} & Contrastive Learning & Swin Transformer & 280 & 768 & TCGA + PAIP & 16 & 32 & no \\ \hline
    \textbf{REMEDIS} & Contrastive Learning & \begin{tabular}[c]{@{}c@{}}BiT-M\\ (ResNet-152x2)\end{tabular} & 240 & 4096 & TCGA & 50 & 29 & no \\ \hline
    \textbf{\begin{tabular}[c]{@{}c@{}}BYOL\\ (TCGA)\end{tabular}} & Self-distillation & ResNet50 & 26 & 2048 & TCGA & 2 & 30 & yes \\ \hline
    \textbf{\begin{tabular}[c]{@{}c@{}}BYOL\\ (CHC)\end{tabular}} & Self-distillation & ResNet50 & 26 & 2048 & \begin{tabular}[c]{@{}c@{}}DLBCL, Herohe,\\ Camelyon16\end{tabular} & 2 & 1 & yes \\ \hline
    \end{tabular}
    }
    }
    \begin{tabular}{cc}
        \\ 
    \end{tabular}
\end{table*}

%% file: figures/supp_results_heatmap_AP.tex
\begin{figure*}
    \centering
    \includegraphics[width=\textwidth]{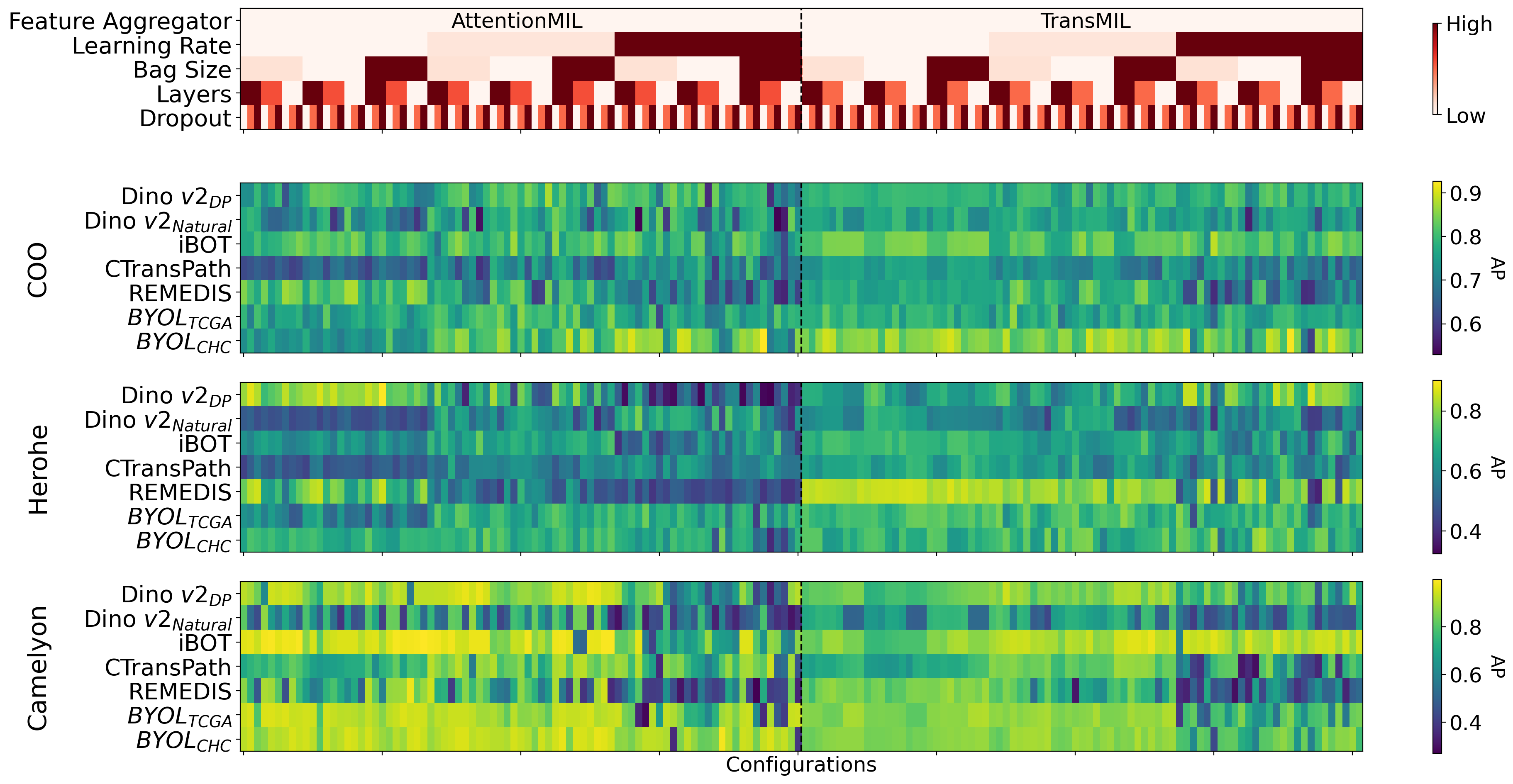}
    \caption{The heatmap shows the performance of every aggregation model configuration for each feature extraction model according to the AP metric. The lighter the color, the higher the AP. The heatmaps are shown for all three datasets. The red-colored legend shows how the aggregation model configurations are ordered on the heatmap. For example, the lowest learning rate is used for the first third of the configurations for each feature aggregator, followed by the next higher learning rate. 
    }
    \label{fig:supp_results_heatmap_AP}
\end{figure*}

%% file: tables/tab_results_supp.tex
\begin{table*}[h]
\centering
\caption{Results for the different feature extractors across all configurations aggregated as mean $\pm$ standard deviation. }
\label{tab:results_mean_std}
\resizebox{\linewidth}{!}{%
\begin{tabular}{|l|r@{}l:r@{}l|r@{}l:r@{}l|r@{}l:r@{}l|}
\hline
\multicolumn{1}{|c|}{\multirow{2}{*}{\makecell{Feature\\extractor}}} & \multicolumn{4}{c|}{COO} & \multicolumn{4}{c|}{Herohe} & \multicolumn{4}{c|}{Camelyon} \\
\cdashline{2-13}
 & \multicolumn{2}{c:}{AUROC} & \multicolumn{2}{c|}{AP} & \multicolumn{2}{c:}{AUROC} & \multicolumn{2}{c|}{AP} & \multicolumn{2}{c:}{AUROC} & \multicolumn{2}{c|}{AP} \\
\hline
$\text{DINOv2}_{\text{DP}}$ & 0.73$\pm$&0.05 & 0.78$\pm$&0.05 & 0.73$\pm$&0.12 & 0.66$\pm$&0.14 & 0.82$\pm$&0.11 & 0.82$\pm$&0.13 \\
$\text{DINOv2}_{\text{natural}}$ & 0.67$\pm$&0.06 & 0.74$\pm$&0.06 & 0.67$\pm$&0.07 & 0.58$\pm$&0.09 & 0.61$\pm$&0.12 & 0.61$\pm$&0.15 \\
iBOT & 0.76$\pm$&0.05 & 0.79$\pm$&0.05 & 0.72$\pm$&0.06 & 0.61$\pm$&0.07 & 0.88$\pm$&0.13 & 0.88$\pm$&0.14 \\
CTransPath & 0.63$\pm$&0.06 & 0.71$\pm$&0.05 & 0.70$\pm$&0.06 & 0.59$\pm$&0.07 & 0.76$\pm$&0.12 & 0.75$\pm$&0.14 \\
REMEDIS & 0.71$\pm$&0.08 & 0.75$\pm$&0.07 & 0.76$\pm$&0.11 & 0.68$\pm$&0.16 & 0.70$\pm$&0.17 & 0.67$\pm$&0.19 \\
$\text{BYOL}_{\text{TCGA}}$ & 0.71$\pm$&0.04 & 0.77$\pm$&0.04 & 0.77$\pm$&0.05 & 0.68$\pm$&0.08 & 0.84$\pm$&0.12 & 0.84$\pm$&0.13 \\
$\text{BYOL}_{\text{CHC}}$ & 0.77$\pm$&0.07 & 0.82$\pm$&0.06 & 0.79$\pm$&0.05 & 0.69$\pm$&0.07 & 0.87$\pm$&0.09 & 0.87$\pm$&0.09 \\
\hline
\end{tabular}
}
\end{table*}